\documentclass[11pt]{article}
\usepackage{graphicx}
\usepackage{amssymb}
\usepackage{amscd}
\usepackage{mathrsfs}
\usepackage{longtable,lscape}
\usepackage{amsthm}
\usepackage{amsfonts}
\usepackage{amsmath}
\usepackage{bbm}
\usepackage{float}
\usepackage{url}

\newcommand{\captionfonts}{\footnotesize}
\makeatletter  
\long\def\@makecaption#1#2{%
  \vskip\abovecaptionskip
  \sbox\@tempboxa{{\captionfonts #1: #2}}%
  \ifdim \wd\@tempboxa >\hsize
    {\captionfonts #1: #2\par}
  \else
    \hbox to\hsize{\hfil\box\@tempboxa\hfil}%
  \fi
  \vskip\belowcaptionskip}
\makeatother 
\begin{document}
\title{The Heart of an Image: Quantum Superposition \\ and Entanglement in Visual Perception}
\author{Jonito Aerts Argu\"elles
 \vspace{0.5 cm} \\ 
        \normalsize\itshape
        KASK and Conservatory, \\
        \normalsize\itshape
         Jozef Kluyskensstraat 2, 9000 Ghent, Belgium
        \\
        \normalsize
        E-Mail: \url{jonitoarguelles@gmail.com}
       	\\
              }
\date{}
\maketitle
\begin{abstract}
\noindent 
We analyse the way in which the principle that
`the whole is greater than the sum of its parts' 
manifests itself with phenomena of visual perception. For this investigation we use insights and techniques coming from quantum cognition, and more specifically we are inspired by the correspondence of this principle with the phenomenon of the conjunction effect in human cognition. We identify entities of meaning within artefacts of visual perception and rely on how such entities are modelled for corpuses of texts such as the webpages of the World-Wide Web for our study of how they appear in phenomena of visual perception. We identify concretely the conjunction effect in visual artefacts and analyse its structure in the example of a photograph. 
We also analyse quantum entanglement between different aspects of meaning in artefacts of visual perception. We confirm its presence by showing that well elected experiments on images retrieved accordingly by Google Images give rise to probabilities and expectation values violating the Clauser Horne Shimony Holt version of Bell's inequalities. We point out how this approach can lead to a mathematical description of the meaning content of a visual artefact such as a photograph.
\end{abstract}
\medskip
{\bf Keywords:} Quantum theory; 
Quantum cognition; 
Visual perception; Conjunction effect; Quantum entanglement; Bell's inequalities.

\section{Introduction\label{intro}}
I have a long-standing interest in ephemeral and emergent phenomena of visual perception.  From how the zoetrope produces the movement out of still images and hence led to the birth of cinema to the way in which holography makes emerge a three-dimensional view from a laser and a two-dimensional photographic plate. The end-work for my Master study in visual art was focused on the phenomenon of the afterimage \cite{aertsarguelles2017a}. This phenomenon takes place on the interface of the interaction between observer, the one who looks, and object, that which is looked at, in a way so that they produce `emergences' 
-- the zoetrope leads in an unexpected way from still images to `movement', holography gives rise to a third dimension from a picture that is two-dimensional, afterimages both positive and negative are self-emergent. I can date back my fascination for these emergent phenomena of visual perception to my original passion for visual art in general and photography specifically. Indeed, even with the simple action of a photograph, it is always the elusive nature of that moment that continues to intrigue me. A photograph is a snapshot of reality, but it is at the same time something much deeper and more meaningful than that moment in itself. As photographer, one never sees the exact image that is recorded. The moment one presses the shutter, the mirror folds up and one sees no longer anything through the camera. 

When photography was introduced one hundred and fifty years ago, there was a tendency to consider it as a perfect medium to document reality. This was because the mechanical quality of the process ensured an unspoiled replica of the subject. The technical progress of the following decades and the rise of photojournalism made for an even clearer and more real looking product. But 
of course, it is not a duplicate of reality at all. Photographers are now able to edit their photographs digitally, although we should not forget that in the time of analogue film editing was already frequent, albeit less sophisticated. At that time, for example, there were 
various ways of developing film and adjustments could be made when printing images. In this way, an image can always be interpreted as the `look of the photographer'. He chooses at least partly what we see, and also what gets the most attention in the image. The image therefore carries more in it than just the literal representation of reality. There is more going on when looking at an image than pure registration. It is this elusive `more' that fascinated me so much already as a boy when I was making my first pictures.

This fascination has recently been triggered to a new level by an experience of scientific collaboration within the framework of the Center Leo Apostel of the Free University of Brussels with a group of mainly physicists, but also deeply involved in interdisciplinary applications of the natural sciences. Not so long ago, one of these physicists published an article entitled `Ephemeral properties and the illusion of microscopic particles' \cite{sassolidebianchi2011}. The article shows how some of the properties of quantum particles can be considered to be ephemeral. The history of quantum mechanics is indeed intertwined with a search for the peculiar nature of quantum entities and their properties \cite{heisenberg1969,einsteinpodolskyrosen1935,bell1987,feynman1965}. There is a consensus among the 
physics' 
experts on the `elusiveness of this peculiarity', 
and more generally 
`the incomprehensibility of the behaviour of quantum entities themselves'. One among the famous ones, Richard Feynman, once expressed such `incomprehensibility and vagueness of the mysterious quantum world' as follows: ``I think I can safely say that nobody understands quantum mechanics."  

It would lead us too far to even give an overview of this situation of the ephemeral and the elusive in the broad field of quantum mechanics. However, the collaboration in which I became involved was focused on a specific sub-research field that is called `quantum cognition'. Strangely enough, this sub-study sheds light on a specific aspect 
of the quantum nature, 
directly linked to the collection of phenomena of human visual perception in which I was involved. 
I mentioned already the notion of emergence and the first quantum effect I would like to consider is related to this phenomenon.

\section{The 
non-summability 
of visual perception}

Let me first clarify what the title of this section refers to.
It specifies the general truth that `the whole is greater than the sum of its parts' 
(hence the term ``non-summability''), 
except in the domain where classical set theory applies. Indeed, if we consider two sets
\begin{eqnarray}
A=\{\rm\it Cat,Horse,Donkey\}\quad\quad\ B=\{\rm\it Cat,Mouse,Squirrel\}
\end{eqnarray}
then these two sets form a whole that in mathematics is called `their union'
\begin{eqnarray}
A\cup B=\{\rm\it Cat,Horse,Donkey,Mouse,Squirrel\}
\end{eqnarray}
which is simply the sum of the two original sets as parts of such whole. One could argue that the two parts contain each three elements while the union contains only five elements, indeed, 
\emph{Cat} 
is present in both parts, whereas it only appears once as an element of their union. In this sense, the use of the word `sum' is slightly misleading, which is why `union' has been introduced as a concept in set theory. The expression that `the whole is greater than the sum (or union) of its parts' means that `by the coming together of the parts, something new is created that makes an essential contribution to the whole'. 

Let me also express the dual aspect of `the whole is greater than the sum of its parts', because the examples we will later deal with 
mainly refer to it. For the sets $A$ and $B$ we can see that the intersection
\begin{eqnarray}
A \cap B=\{\rm\it Cat\}
\end{eqnarray}
is always smaller than each of the $A$ and $B$ collections separately. This set-theoretical law that `the intersection is always lesser than each of its elements' can also lead to problems in those domains where classical set theory does not apply, where an intersection can behave as if it was larger than one of its elements, or the other. As I said, this is the dual version of the phenomenon where the whole appears to be larger than the sum of its parts.
In the course of the last century, it became clear during a continuous in-depth investigation into the structure of quantum mechanics that the latter carries inherently in its principles this `whole being greater than the sum of its parts' aspect. More specifically, if two entities $A$ and $B$ are brought together, then the new composite entity $C$, as the whole formed from parts $A$ and $B$, is different (larger) than their sum (union), so we can write
\begin{eqnarray}
C>A\cup B
\end{eqnarray}
It is a well-defined principle in quantum mechanics, called `the superposition principle', which is at the origin of this phenomenon of the whole being greater than the sum of its parts. A clear analysis of this `quantum phenomenon of magnification of the whole after composition of its parts', which goes into all the necessary details, has only been recently established. 
However, it was already taken into due account in the early years of quantum mechanics, although rather at an intuitive level, as evidenced by the book written by Werner Heisenberg, one of the founders of quantum mechanics, with the title: ``The part and the whole" \cite{heisenberg1969}.  In quantum cognition, and therefore also in the research in which I participated at the Center Leo Apostel, effective use is being made of this specific quantum feature, to build models from the mathematics of quantum mechanics for still unknown phenomena of human cognition. Let me look at some of these phenomena, so that I can use them to make it clear what the possible applications are for the field of visual perception.

\section{The Linda problem}

The first example I will consider is in the field of human cognition, and it has already been extensively dealt with in quantum cognition approaches. In the 1990s, psychologists Amos Tversky and Daniel 
Kahneman 
discovered and identified the cognitive phenomenon called the `conjunction fallacy' \cite{tverskykahneman1982}.  A well-known example of the conjunction fallacy, which has meanwhile been considered archetypical, is called the `Linda problem'. It is 
about asking 
the following: 
\begin{quote}
``Linda is 31 years old, single, outspoken, and very bright. She majored in philosophy. As a student, she was deeply concerned with issues of discrimination and social justice, and also participated in anti-nuclear demonstrations. Which is more probable? $(A)$ Linda is a bank teller. $(B)$ Linda is a bank teller and is active in the feminist movement."
\end{quote}
The majority of  
individuals subjected to the above 
`Linda problem' 
alternative 
opt for option 
$(B)$. 
However, the probability that two events occur together, i.e. in `conjunction', is always less than or equal to the probability that one of the two occur alone. Indeed, for the two events 
$(A)$ and $(B)$, 
this can be written as the two inequalities:
\begin{eqnarray}
P(A\ {\rm and}\ B) \le P(A)\quad\quad P(A\ {\rm and}\ B) \le P(B)
\end{eqnarray}
whereas people who are presented with the Linda problem will generally claim that
\begin{eqnarray}
P(A\ {\rm and}\ B)> P(A)
\end{eqnarray}
Let us take a brief look at some concrete values for these probabilities, to see what `error' is being made. Suppose there is only a very small probability that Linda is a bank teller, for example 
$P (A)=0.05$
and a high probability that Linda is feminist, for example 
$P(B)=0.95$, 
then the probability that Linda is both must anyhow be smaller than each of the two separately, and if they are independent of one another, this probability must be equal to the product $0.05\times 0.95=0.0475$, and this is smaller than $0.95$ and smaller than $0.05$.
What happens if an overwhelming majority of those questioned assesses these probabilities in such a wrong way? The classical answer is that the interviewees make a `probability judgment error', hence the name `conjunction fallacy' given by Tversky and Kahneman to the phenomenon. To measure the importance of the phenomenon, we note that the `conjunction fallacy' occurs very often in stories like the one about Linda, and is therefore not at all the result of a deliberate `deception' of the interviewees \cite{moro2009}.
In quantum cognition, a totally different explanation is put forward. It is not assumed that the interviewees make a mistake in their assessment, on the contrary, they are considered to be `correctly judging' what the probabilities in question are. 

In order to be able to clarify the basis of this totally different and new way of considering this phenomenon of `conjunction fallacy', from the perspective of quantum cognition, it will be useful to first consider a second example, which I myself have been involved with when working with the group of researchers at the Center Leo Apostel of the VUB. This second example, situated in the field of concept theory research in psychology, shows on a deeper level what is happening in the formation of a `conjunction that seems greater than one of its elements', as is the case for the Linda problem. After all, there too, a decade before the conjunction fallacy was identified by Tversky and Kahneman, a peculiarity was noted in connection with the conjunction of two concepts, which was called `the pet-fish problem'.

\section{The pet-fish problem}

In 1981, Daniel Osherson and Edward Smith studied the combination of the two concepts {\it Pet} and {\it Fish} in the conjunction {\it Pet-Fish} and 
considered `how typical' different examples of the concepts of {\it Pet}, {\it Fish} and {\it Pet-Fish} are estimated 
\cite{oshersonsmith1981}. They noticed that examples such as {\it Guppy} and {\it Goldfish} are much more typical of the 
{\it Pet-Fish} 
conjunction than is the case for the separate concepts of {\it Pet} and {\it Fish}, and showed that this is in contradiction with a modelling of the situation by classical fuzzy set theory. 
Indeed, as 
is the case for 
the 
probabilities in the Linda problem, fuzzy set
theory also demands that the typicality of an example with respect to the conjunction of two concepts is less or equal to the typicality of that example for each of the concepts taken separately. 
And since 
{\it Guppy} was the item that Osherson and Smith put forward with respect to the pet-fish problem, the phenomenon is also often called the `guppy effect'. 

What is important, and in the early stages of the investigation was not clear, is that 
this 
guppy effect is very often present in combinations of two concepts. For example, studies have shown that respondents believe that {\it Olive} is a more typical example of 
{\it Fruits and Vegetables} 
than of {\it Fruits}, or {\it Vegetables}, considered separately 
(see Table 4a of \cite{sozzo2015}), and this is also the case for {\it Mint} with respect to {\it Food and Plant} and {\it Tree House} with respect to {\it Building and Dwelling} (see \cite{hampton1988} and Table 4 of \cite{aerts2009a}).
Not so 
however for {\it Apple}, which is found to be more typical of {\it Fruits} than of 
{\it Fruits and Vegetables}, 
or of {\it Vegetables}, while respondents find {\it Broccoli} to be a more typical example of {\it Vegetables} than of 
{\it Fruits and Vegetables}, 
or of {\it Fruits}. But even {\it Apple} is found to be more typical of {\it Fruits and Vegetables} than of {\it Vegetables} and {\it Broccoli} is found to be more typical of {\it Fruits and Vegetables} than of {\it Fruits} (see Table 4a of \cite{sozzo2015}). 
Psychologists who study concepts and their conjunctions \cite{hampton1988}, called `overextension'
the effect where an exemplar (such as {\it Apple}) is more typical for the conjunction of two concepts (such as {\it Fruits and Vegetables}) as for one of the concepts apart (such as {\it Vegetables}), and found it 
to be 
a very frequent phenomenon (see Tables 1a, 2a, 3a and 4a of \cite{sozzo2015} and Table 4 of \cite{aerts2009a}), so certainly not a rare phenomenon as Osherson and Smith could still think back in 1981. 
The exemplars {\it Olive}, {\it Mint} and {\it Tree House} are `double overextended' respectively with respect to the concepts combinations {\it Fruits and Vegetables}, {\it Food and Plant} and {\it Building and Dwelling}, hence they are found to be more typical with respect to the conjunction as with respect to both concepts aparts. Such a double overextension, although it occurs systematically, is not abundant like it is the case for a single overextension.
By analyzing 
the examples of overextension, and more even so for the examples of double overextension, in detail, we can understand much better what is going on and why quantum cognition is able to bring forward a fundamental solution for the `conjunction phenomenon in general', namely that `the conjunction can be perceived to be larger than its composing elements'.

In order to understand what is going on, let us consider the conjunction 
{\it Fruits and Vegetables}. 
Examples that show overextension, and certainly so the ones that show double overextension, contain a typicality of what the conjunction 
introduces at the level of meaning itself, 
let us illustrate this by means of the example {\it Olive}.
For an olive, we can say that there are doubts whether it is a fruit or a vegetable, and therefore, because there is such doubt, 
an olive is more typical of 
{\it Fruits and Vegetables}  
than of {\it Fruits}, or of {\it Vegetables}. It is as if 
{\it Olive} was located in an area where {\it Fruits} and {\it Vegetables} 
introduce 
something that is `new', also from a structural viewpoint, which is neither in the area of {\it Fruits} nor in the area of {\it Vegetables}, so not in what we call the traditional conjunction. This `new' aspect 
manifests 
as a consequence of `the specific meaning of the conjunction' which allows to bring doubtful cases into its realm, while the meaning of the conjunction based on set theory and hence logic `does not allow such doubtful cases'. From the viewpoint of the latter, something can  be part of it only if it is in `both', i.e. 
part of {\it Fruits} and at the same time part of {\it Vegetables}, 
and not if it is a situation of doubt, 
i.e. perhaps part of {\it Fruits} or perhaps part of {\it Vegetables}. 
That is precisely what is allowed in quantum cognition. 
This is also the case for {\it Mint} and {\it Tree House}, they give rise to such a doubt, respectively with respect to being a plant or food, and being a building or a dwelling, and are therefore located in this new area, which does not exist within classical set theory. 

Since fuzzy set theory is still set theory from a mathematical point of view, a modelling that starts from it will still not be sufficient to take into due account all these situations of doubt. On the other hand, since quantum cognition uses the mathematical formalism of quantum mechanics, it becomes possible to model this phenomenon of 
`overextension of the probabilities'. 
We can make this clear without using the difficult and sophisticated mathematics of quantum 
mechanics, 
by directly considering the underlying structures of both classical and quantum 
theories, which is what we are now going to do.

\section{Classical and quantum}

\begin{figure}[H]
\begin{center}
\includegraphics[height=4cm]{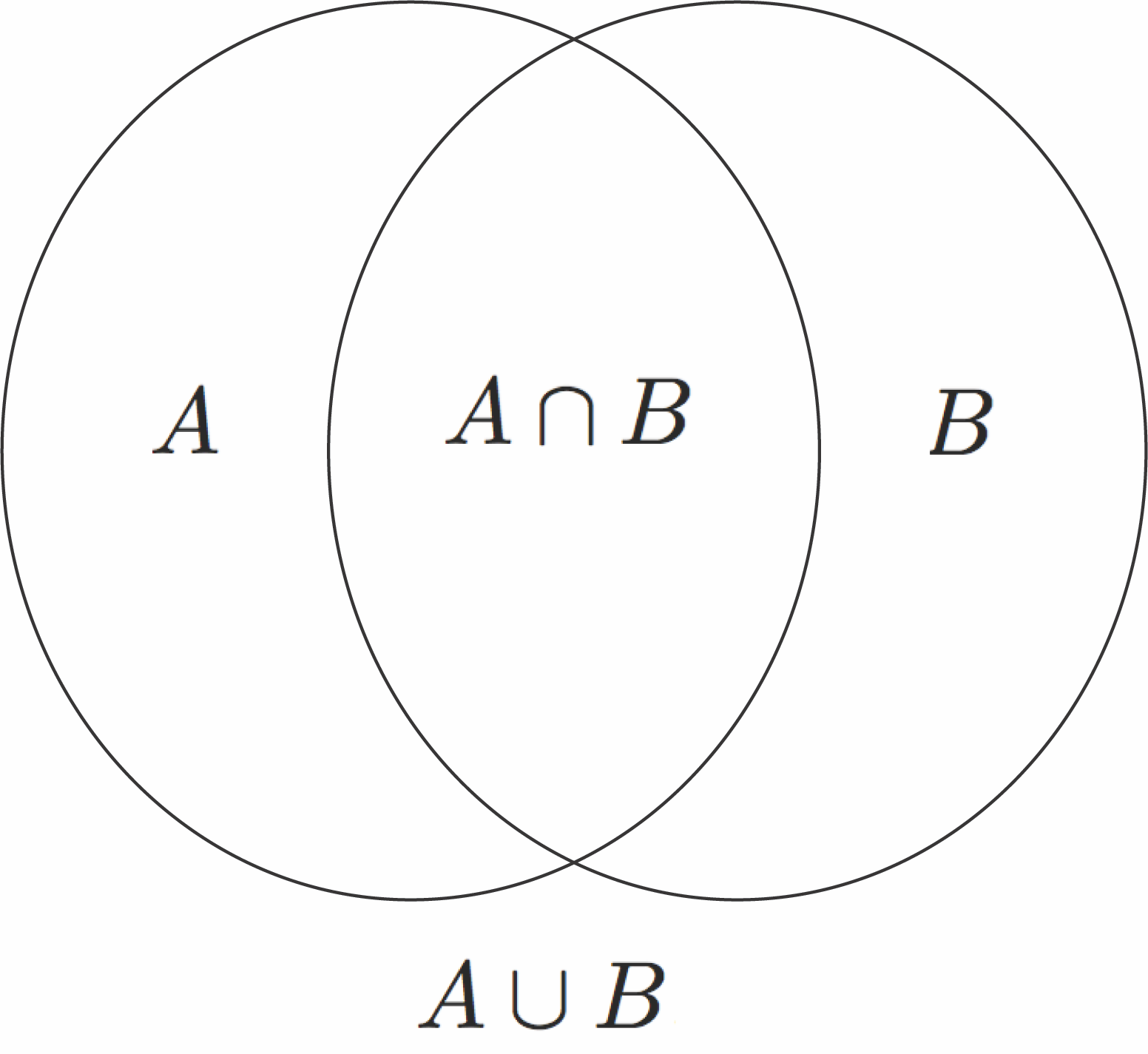}
\caption{A graphical representation of two sets $A$ and $B$ in the plane, and 
of 
their intersection $A \cap B$ and 
union $A \cup B$. If 
$\mu$ is the `Lebesgue measure', coinciding here with the standard measure of the area,
we can read from the drawing that 
$\mu(A \cup B) \le \mu(A) + \mu(B)$, 
$\mu(A \cap B) \le \mu(A)$ and $\mu(A \cap B) \le \mu(B)$.
\label{unionintersection}}
\end{center}
\end{figure}

We already noted that classical theories are based on set theory, so probabilities and measurements of typicalities will necessarily be mathematically expressed by what is called a 
`measure on sets'.
Such a 
measure, 
even in the most general case, behaves like the measure of a surface, if we can imagine these sets to be portions of a plane. We typically have the situation described in Figure~\ref{unionintersection}, where two sets $A$ and $B$ are represented, with their intersection $A \cap B$ and union $A \cup B$.
It is clear that the 
area of $A \cap B$ is always smaller or equal to the sum of the 
area of $A$ and the 
area of $B$:
\begin{eqnarray}
\mu(A \cup B) \le \mu(A) + \mu(B)
\end{eqnarray} 
This shows that the principle `the whole $A \cup B$ is greater than the sum of its parts $A$ and $B$' never takes place when the modelling is done from set theory. What we can also easily see from Figure~\ref{unionintersection} is that the 
area of $A \cap B$ is always lesser or equal to both the 
area of $A$ and 
the area 
of $B$:
\begin{eqnarray}
\mu(A \cap B) \le \mu(A)\quad\quad \mu(A \cap B) \le \mu(B)
\end{eqnarray}
This shows that what happens with the `conjunction effect', both in the case of the Linda problem and in the case of the pet-fish problem, never takes place when the modelling is done from set theory. 

\begin{figure}[H]
\begin{center}
\includegraphics[height=4cm]{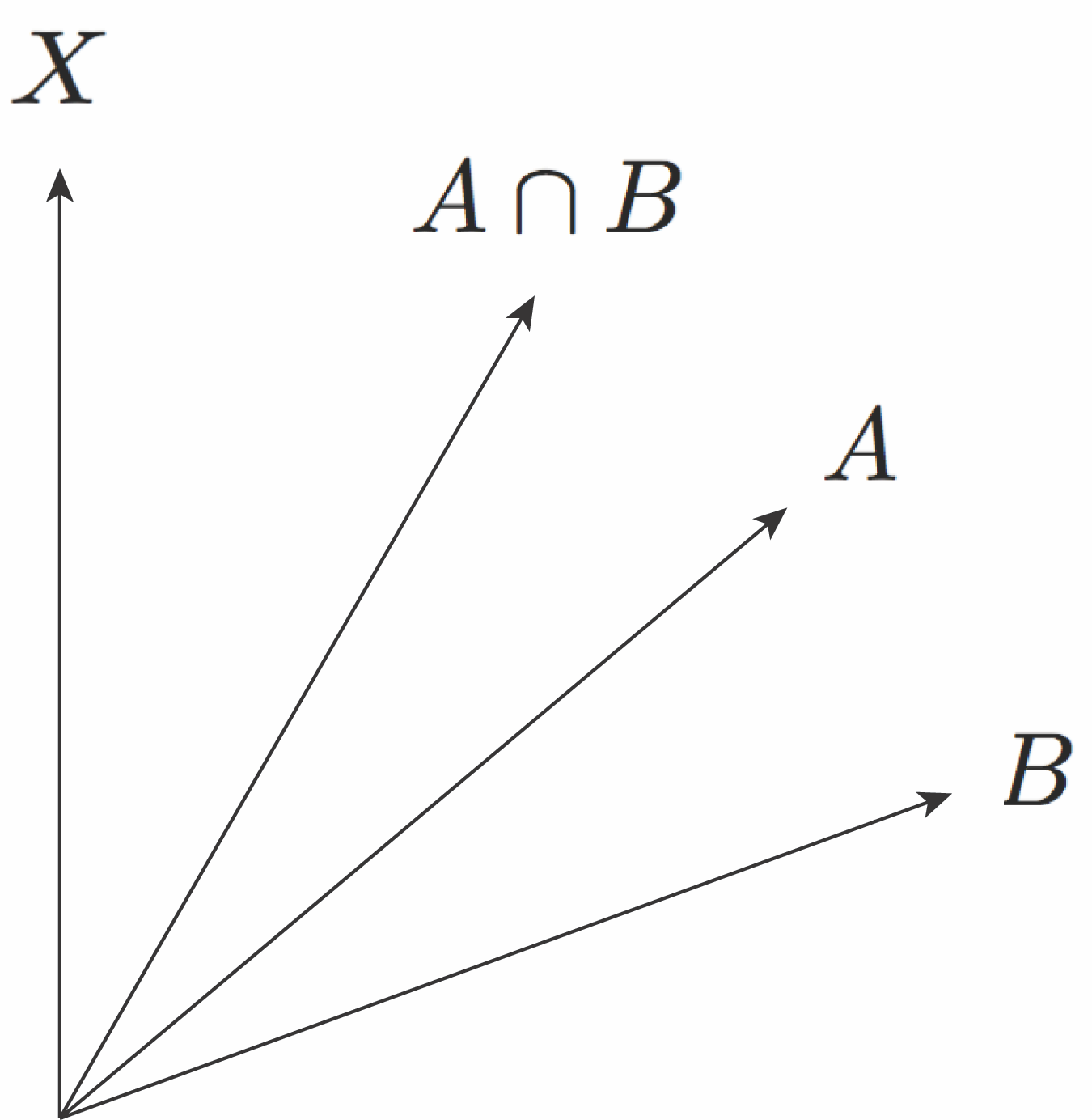}
\caption{A graphical representation of four vectors $X$, $A \cap B$, $A$ and 
$B$, 
of a Hilbert space representing the situation of a quantum model able to model the conjunction phenomenon, i.e. 
the inequalities 
$\mu(A \cap B) > \mu(A)$ and $\mu(A \cap B) > \mu(B)$.
\label{quantummodel}}
\end{center}
\end{figure}

What is the underlying structure of quantum mathematics that does make it possible to model the `conjunction phenomenon'? Well, `state spaces' in quantum mechanics are described by `Hilbert spaces', which are 
`vector spaces over the complex numbers', possibly infinitely dimensional.
\noindent
To be able to understand why the phenomenon of `the whole that is larger than the sum of its parts' and of `the conjunction that is 
larger than its elements', can be described by means of a quantum mathematical model, it will be enough to  
consider a simplified 
version of such model, and more precisely a two-dimensional  
real 
version that can be described on a plane. For this, let us consider four vectors lying on a same 
plane, which 
we call $A$, $B$, $C$ and $X$, as shown in Figure~\ref{quantummodel}. 
\noindent
In the case of the Linda problem, $A$ represents `bank teller', $B$ represents `feminist', and $C$ represents `bank teller and feminist', while $X$ represents `Linda and her description'. For the 
pet-fish 
problem, $A$ represents {\it Pet}, $B$ represents {\it Fish}, and $C$ represents {\it Pet-Fish}, while $X$ represents {\it Guppy}.

Now, not 
only in quantum mechanics properties are represented by subspaces of 
the Hilbert space (which in our example are the one-dimensional subspaces generated by the different vectors on the plane),   
 rather than by sets, but probabilities are also defined in a fundamentally different way. With the exception of the
more 
elaborate quantum models developed 
following the `hidden-measurement approach', 
where different 
`measurement-interactions' are assumed to play a role in the selection of an outcome 
\cite{aertssassolidebianchi2015a,aertssassolidebianchi2015b}, the quantum probabilities in standard quantum theory are not linked to a 
direct 
measure of properties or states, but to the 
angles 
that exist between subspaces of state vectors. It is not necessary here to specify 
the 
mathematical formula 
exactly used 
in 
our demonstration 
(the so-called Born rule), being 
enough to know, in our simple example on the plane, that if the angle between two vectors is $\theta = 90^\circ$, which means that the two vectors are perpendicular to each other, then the probability associated with them is $P = 0$. If the angle is $\theta = 0^\circ$, i.e. if the vectors have the same direction, then the probability associated with them is equal to $P = 1$. 
Also, when 
the angle decreases, from $\theta = 90^\circ$ to $\theta = 0^\circ$, the corresponding probability progressively increases, from $P = 0$ to $P = 1$, 
like the square of the cosine of the angle.

\section{Quantum modeling the Linda and pet-fish problems}

\begin{figure}[H]
\begin{center}
\includegraphics[height=4cm]{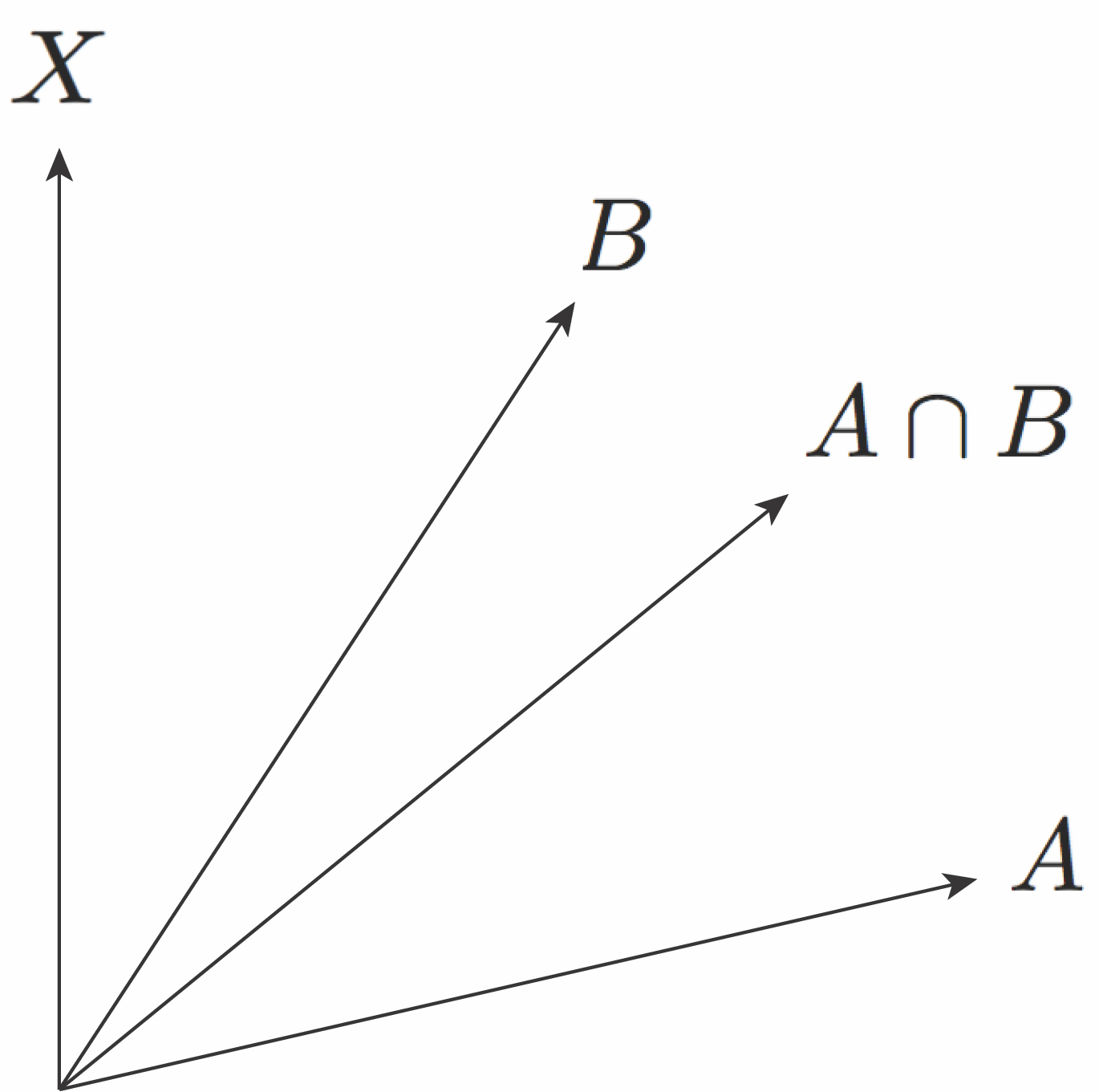}
\caption{A graphical representation of four vectors $X$, $A \cap B$, $A$ and $B$, 
providing 
a quantum model for the data corresponding to the Linda, 
where 
$\theta(X,A)=33.21^\circ$, 
$\theta(X,A \cap B)=50.77^\circ$ and $\theta(X,B)=77.08^\circ$.
\label{quantummodellinda}}
\end{center}
\end{figure}

\begin{figure}[H]
\begin{center}
\includegraphics[height=4cm]{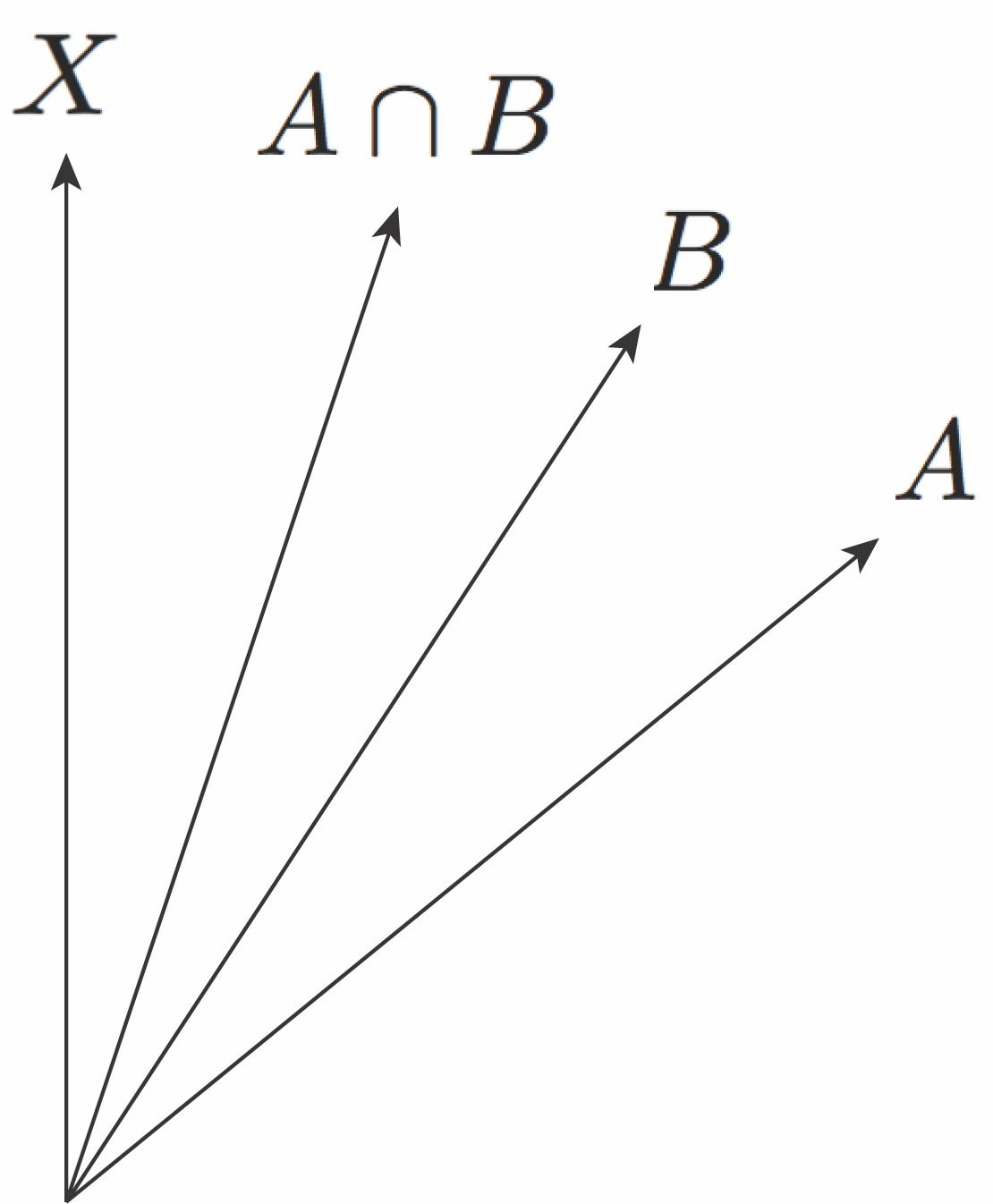}
\caption{A graphical representation of four vectors $X$, $A \cap B$, $A$ and $B$, 
providing 
a quantum model for the data corresponding to the pet-fish problem, 
where 
$\theta(X,A \cap B)=18.43^\circ$, $\theta(X,B)=33.21^\circ$ and $\theta(X,A)=50.77^\circ$.
\label{quantummodelpet-fish}}
\end{center}
\end{figure}

To be as specific as possible, let us assume that for the Linda problem the probability that Linda is a `bank teller', a `bank teller and feminist' and a 
`feminist', 
were estimated by the interviewees to be $P_{\rm\it Linda}(A)=0.05$, $P_{\rm\it  Linda}(A\cap B)=0.4$ and $P_{\rm\it  Linda}(B)=0.7$, respectively. For the pet-fish problem, we assume that the degree of typicality of {\it Guppy} for {\it Pet}, {\it Pet-Fish} and {\it Fish}, were estimated to be 
$P_{\rm\it  Pet-Fish}(A)=0.4$, $P_{\rm\it  Pet-Fish} (A \cap B)=0.9$ and $P_{\rm\it  Pet-Fish}(B)=0.7$, 
respectively. In Figures~\ref{quantummodellinda} and \ref{quantummodelpet-fish} we have a depiction of the quantum models that can fit these data using a simple quantum vector space on the 
real 
plane. 
\noindent
Indeed, in Figure~\ref{quantummodellinda} the angle between $X$ and $B$, $\theta(X,B)$, is the smallest, so corresponding with the highest probability 
$P_{\rm\it  Linda}(A)=0.7$, 
and using the 
quantum mechanical (Born) rule of probabilistic assignment 
we have: 
\begin{eqnarray}
P(A)=\cos^2\theta(X,A) \quad \Leftrightarrow \quad \theta(X,A)= \arccos\sqrt{P(A)} \pmod{180^\circ}
\label{Born}
\end{eqnarray}
so that 
we find 
the angles (modulo $180^\circ$): 
$\theta(X,A)=33.21^\circ$,
$\theta(X,A \cap B)=50.77^\circ$ and $\theta(X,A)=77.08^\circ$.  
Similarly, for the pet-fish problem, 
Figure~\ref{quantummodelpet-fish} shows that the angle between $X$ and $A \cap B$, $\theta(X,A \cap B)$, is the smallest, so corresponding 
to 
the highest probability $P_{\rm\it Pet-Fish}(A \cap B)=0.9$,  
so (\ref{Born})  gives 
$\theta(X,A \cap B)=18.43^\circ$, 
and we also obtain  
$\theta(X,B)=33.21^\circ$ and $\theta(X,A)=50.77^\circ$.

While it is true that more elaborated models are used in quantum cognition, with higher dimensional vector spaces and corresponding subspaces, it is already possible to see in this very simple two-dimensional example how the limitations that classical models carry with them (as a consequence of being formulated by means of set theory) can be easily overcome, using a probability calculus based on vectors spaces.
 
The conjunction fallacy, in the form identified by Tversky and Kahneman, was studied in quantum cognition by the group of Jerome Busemeyer at Indiana University \cite{busemeyerpothosfrancotrueblood2011,truebloodbusemeyer2011} and also as one of many other related phenomena of a non-classical nature that can occur in human decision-making processes, by the group of Emanuel Pothos at the City University of London, at the International Centre of Mathematical Modeling of Linnaeus University in Sweden (Andrei Khrennikov), and in the School of Business of Leicester University (Emanuel Haven) \cite{truebloodbusemeyer2011,busemeyerbruza2012,pothosbusemeyer2009,pothosbusemeyer2013,khrennikov2010,havenkhrennikox2013}. The research into the way in which quantum models can describe concepts and their combinations, including the conjunction of concepts, was largely undertaken by the group at the Center Leo Apostel of 
VUB, in collaboration with the Universities of British Columbia, Leicester and Gdansk \cite{aerts2009a,aertsgabora2005a,aertsgabora2005b,aertsgaborasozzo2013,aertsczachor2004,aertssozzo2011,aertsbroekaertgabora2011,gaborakitto2016,sozzo2014,aertsetal2017a,aertsetal2017b,aertsetal2017c,aertsetal2017d,aertsetal2017e}. In particular, the research that led to the identification of the conjunction effect in texts of the World-Wide Web \cite{aertsetal2017b} has been important 
to find a way to 
identify and investigate the conjunction effect in visual perception,  
as I'm going to describe in the following section, 
and has also led to a quantum model for the World-Wide Web itself, which we called the QWeb \cite{aertsetal2017e}.

\section{The conjunction effect in visual perception}

For the development of our quantum model for the World-Wide Web, we showed 
in \cite{aertsetal2017e} 
how a quantum-like `entity of meaning' can be linked to a `language entity consisting of printed documents', where we considered the latter as a collection of traces left behind by the former, in specific results of searches we 
described 
as measurements. In other words, we offered a perspective where a collection of documents such as the World-Wide Web is described as a space of 
manifestation 
of a more complex entity -- the QWeb -- that was the object of our modelling and for which we used inspiration and experience coming from previous studies 
on 
operational-realistic approaches to quantum mechanics and quantum modelling of human cognition and decision processes. We 
also emphasized in \cite{aertsetal2017e} 
that a consistent QWeb model must account for the observed correlations between words found in the printed documents, such as `co-occurrences', since the latter are connected to `meaning connections' that exist between the concepts associated with these words. In this respect, we 
showed 
that `context and interference (quantum) effects' are necessary to explain the probabilities calculated by counting the relative numbers of documents containing specific words and the co-occurrence of 
these 
words \cite{aertsetal2017e}.

\begin{figure}[H]
\begin{center}
\includegraphics[height=8cm]{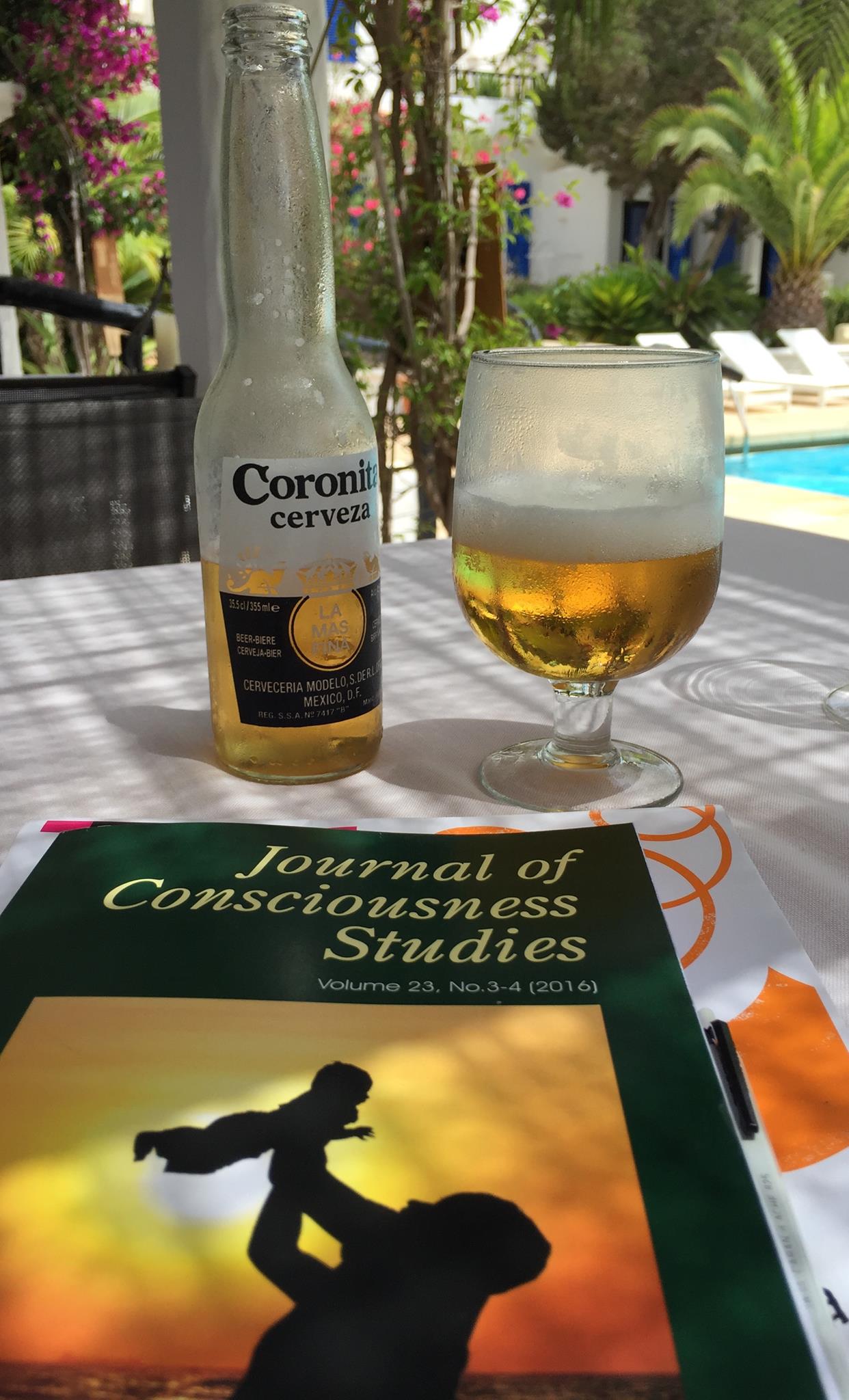}
\caption{A photograph with different examples of conjunction effects (courtesy of Massimiliano Sassoli de Bianchi)
\label{mother-child-bottle-glass}}
\end{center}
\end{figure}

It is from the insight we gained in the study of texts of the World-Wide Web that 
I can also propose here 
a quantitative modelling of the `the meaning that is contained in entities of visual perception'. This general quantum modelling will also be applicable in the specific case where the visual entity is a photograph. Let us  consider the specific photo of Figure~\ref{mother-child-bottle-glass} and use it to explain in some detail the idea behind 
the 
approach. Two striking `conjunctions' can be seen in the photo that produce a `conjunction effect'. The first is `the bottle and the glass' and the second is `the mother and the child'. Let us consider some elements that can make these conjunction effects clearer. Consider the concept of {\it Thirst}. If we consider a photo with only a bottle, or a glass, then we can suspect that {\it Thirst} will be evoked less effectively by looking at it than is the case for this photo, where both the glass and the bottle are present. 
For the conjunction of {\it Mother} and {\it Child} we can consider the element {\it Love}. A photo of a child alone or a mother alone will evoke much less the element of {\it Love} than the connection between mother and child shown in this photo. 
\noindent
But there are also more subtle conjunction effects present in the picture. For example, the conjunction of {\it Mind} and {\it Consciousness}. There is the {\it Spirit} and {\it Bottle} and {\it Consciousness} as a modern name for {\it Spirit}, and the {\it Journal of Consciousness Studies} which is a new entity, whereas the beer in the bottle and glass is a very old alcoholic beverage. There is also a subtler connection between {\it Mother} and {\it Child} on the cover of the 
magazine: 
the depiction of mother and child explicitly shows a `spatial connection', the mother has the child in her arms, and at the same time an emotional connection, the love between mother and child. There is also a connection between the bottle and the glass, on the one hand, spatially, by the action of someone who drinks, by standing on the same table, and also {\it Bottle} and {\it Glass} are emotionally/energetically connected, the bottle empties itself to fill the glass. Each of these elements also contains such conjunctions effects, 
that is, 
effects that indicate that `the whole is greater than the sum of its parts'. By the analysis of this specific photograph it is easy to understand that every photo, and by extension each image, has built-in such conjunction effects, and as a consequence 
the `whole that is greater than the sum of its parts' will be true for each image.

\section{The ontology of an image}

The insight that for each image the phenomenon of `the whole that is greater than the sum of its parts' is present, will not create any surprise in the context of artistic interpretation of visual entities such as images. However, what we would like to put forward here is that from 
the perspective of 
quantum cognition and the associated mathematical models, it becomes possible to investigate this phenomenon quantitatively. Such an approach also allows us to understand in a more precise way what is going on in this respect with a visual entity, such as an image.

In the previous discussion, we used two specific ways to express what is going on that can bring us to the path of a deeper understanding of the conjunction effects in visual entities such as photos. In a first part of our discussion we used the expression 
``...will 
evoke much less the element of {\it Love} than the..." and later we wrote 
``...`contains' such conjunction effects...". 
These 
two verbs, `evoking' and `containing', 
are those 
that we would like to consider now in more detail. The first refers to a `creative event' that is directly connected to the person looking at the photograph, or more generally, the person who interacts with the visual entity. It indicates that there is a creative action of the viewer. The second verb, `contain', refers to the presence, `independently of the creative action of the viewer', of an element of meaning in the photograph itself. Intuitively we know that both are present. It can be shown that quantum mechanics is a theory that describes both aspects in a way such that there is a balance between a `creation aspect', introduced by the viewer, and a `discovery aspect', which is already present and is independent of whether the photograph is looked at or not 
\cite{aertssassolidebianchi2015a,aerts1998,aertscoecke1999}.

Physicists refer to `the existing aspect' that is independent of the observer by means of the notion of `state'. This corresponds to how we also use the notion of `state' in our daily lives. For example, when we say that `the car is in a worn state', we mean `what the car is, regardless of what someone does with it'. Thus, `state' refers to the `ontology' of an entity. The creation aspect that an observer, a viewer when talking about a visual entity, adds to when interacting with an entity in a certain state, physicist will call it a `measurement'. This is a notion that is used mainly in science, and one can feel the bias of science in it, where `interacting with the ontological' corresponds to a measuring process. We will therefore not use the term `measurement' and rather talk more generally about `observing', `perceiving' or `interacting'. However, this does not mean that we cannot make use of the way quantum mechanics describes the act of `measuring', to also describe our `observing', `perceiving' or `interacting'. This is also what is done in quantum cognition when quantum mechanical models are developed 
to describe 
situations in human cognition. 
It 
is also in this way that we plan to use the quantum mechanical models 
to describe 
situations in visual perception, where the `ontology' of the visual entity under consideration is 
accounted for 
by its `state' and the observer's interaction by its `measurement', as quantum mechanics does in the description of physical systems.

What is the ontology of a visual entity, and more specifically, `what is the ontology of a photograph'? Thus, what is the meaning of a photograph that exists `independently' of those who perceive or look at it? This is a question that we would like to examine in more detail now, because we can show how our previous analysis surprisingly provides an answer that would not have been obvious without it. First of all, the ontology of a photo is `not' what is printed on the piece of paper that serves as a carrier for the photo, or is `not' what appears as coloured pixels on a computer or TV screen.  If that would be true, if the ontology of a photograph were to be equal to what is printed or appears on a screen, then `the whole would not be greater than the sum of its parts', because what is printed, or what appears on a screen, is perfectly well described by set theory. It is enough to attribute the different points or pixels to the different sets and the situation presented in Figure~\ref{unionintersection} is then perfectly applicable.

The identification of conjunction effects gives us a hint of what the ontology of a visual entity is: 
it is the place the visual entity occupies in the global realm of human culture. 
Let me 
make this more concrete for the case of a photo. The ontology of a photograph, 
 i.e. 
`what a photograph is', 
regardless of whether it is viewed or not, is 
`the place that the photograph occupies within the global realm of photos in human culture'. 
We can `test' 
this definition of ontology of a photo. For this, let us consider the conjunction 
{\it Mother and child}   
of 
two 
images/concepts  
{\it Mother} and {\it Child}, 
and 
let us 
take 
`Google Images' 
as a representation of the global realm of photos in human culture. If we type 
`mother  and child' into Google Images -- more specifically the three words `mother' `and' and `child', hence we did not force Google to search for the query of the exact combination, which would also have been an option --, we will find somewhat around 500 
pictures showing a mother and a child, 
before others appear, where only a child or a mother are depicted separately. 
On the other hand, if 
we now connect two 
concepts 
that are not `meaning-connected' in our human culture, by means of 
a conjunction, 
e.g. 
{\it Giraffe and Salamander}, 
and we type `giraffe and salamander' in Google Images, we will find practically only photos on which only one of the two are depicted.  This means that {\it Giraffe} and {\it Salamander} are very weakly meaning-connected in our human culture, whereas {\it Mother} and {\it Child} are very strongly meaning-connected. The conjunction effect thus refers to the `meaning presence' of those ontological aspects of a photograph that are `different from what has been printed or what is visible on a screen'. 
One can imagine that where the conjunction effects are present in a photograph, there is something that `protrudes from the paper or emerges from the screen', inhabiting the higher-dimensional space of our global human culture. It is no coincidence that when we define the ontology of entities (in our case visual entities like photographs) in this way, quantum mechanics is able to model them. Indeed, the coherence that we observe in our human culture is very similar, in terms of substance, to the coherence characterizing the quantum entities and their interactions \cite{aerts2009b}.

\section{Violation of the conjunction inequalities}

We can however  
identify, 
also in a real quantitative way, 
the presence of overextensions identical to those 
in the Linda 
and 
pet-fish 
problems, 
by making use of the Google Images search tool. We notice that for a search in Google Images usually around 5 pictures are shown in a row (for a typical screen size) and the number of rows is then around 80, which means that Google Images shows some 400 pictures. Let us first consider the conjunction effect connected to {\it Mother}, {\it Child} and 
{\it Mother and Child}. 
We make a Google 
Images 
search of the term `mother' and keep the page of around 400 pictures open, then we make in another window of our browser a Google Images search of the term `child', and keep also the page of around 400 pictures open, and finally in a third window of our browser we make a Google Images search 
of 
the exact sentence `mother and child' and keep also 
in this case 
these around 400 pictures open. 

As a sign of 
{\it Love}, 
we can look for the presence of an `embrace' 
in 
the different pictures. 
I 
did this on September 21, 2017, and found the following results. The number 
$n({\rm mother})$ 
of pictures open in the `mother' search was 415, and 
$n({\rm embrace, mother}) = 155$ 
showed an `embrace', the number 
$n$(child) 
of pictures open in the `child' search was 475, and 
$n$(embrace, child) = 22 
showed an `embrace', and the number 
$n$(mother and child)  
of pictures open in the `mother and child' search were 365, and 
$n$(embrace, mother and child) = 303   
showed an `embrace'. If we consider the fractions of `embraces' present as a good measure of the presence of {\it Love}, as a meaning, in the considered type of picture, we find
\begin{eqnarray} 
&&P_{Love}({\rm\it Mother}) = {n({\rm embrace},{\rm mother})\over n({\rm mother})}={155 \over 415}\approx 0.37 \\ 
&&P_{Love}({\rm\it Child}) = {n({\rm embrace},{\rm child})\over n({\rm child})}= {22 \over 475}\approx 0.05 \\ 
&&P_{Love}({\rm\it Mother\, and\, Child}) = {n({\rm embrace},{\rm mother\, and\, child})\over n({\rm mother\, and\, child})}= {303 \over 365}\approx 0.83 
\end{eqnarray}
which shows that
\begin{equation}
P_{Love}({\rm\it Mother\, and\, Child})  > P_{Love}({\rm\it Mother}) \quad P_{Love}({\rm\it Mother\, and\, Child}) > P_{Love}({\rm\it Child})
\end{equation}
for the notion {\it Love}, if we consider the presence of an `embrace' as the sign for the notion of {\it Love}.

Let us consider now the concepts {\it Bottle}, {\it Glass} and 
{\it Bottle and Glass}, 
and as a sign for {\it Thirst} we verify whether some `liquid' is present in the considered picture. We again use Google Images in three open browser windows for searches of `bottle', `glass' and `bottle and glass', and count the number of pictures in each open browser window, and each time also the number of pictures where `liquid' is present. This gives us a number of 325 pictures in the search window for `bottle', with 52 showing the presence of `liquid', a number of 390 pictures in the search window for `glass', with 17 with showing the presence of `liquid', and a number of 310 pictures in the search window for `bottle and glass', with 92 showing the presence of `liquid'. If we consider the fractions of those where some `liquid' is present as a good measure of the meaning presence of {\it Thirst} in the considered type of 
pictures, 
we find
\begin{eqnarray} 
&&P_{Thirst}({\rm\it Bottle}) = {n({\rm liquid},{\rm bottle})\over n({\rm bottle})}={52 \over 325}\approx 0.16 \\ 
&&P_{Thirst}({\rm\it Glass}) = {n({\rm liquid},{\rm glass})\over n({\rm glass})}= {17 \over 390}\approx 0.04 \\ 
&&P_{Thirst}({\rm\it Bottle\, and\, Glass}) = {n({\rm liquid},{\rm bottle\, and\, glass})\over n({\rm bottle\, and\, glass})}= {92 \over 310}\approx 0.30
\end{eqnarray}
which shows that
\begin{equation}
P_{{Thirst}}({\rm\it Bottle\, and\, Glass})  > P_{{Thirst}}({\rm\it Bottle}) \quad P_{{Thirst}}({\rm\it Bottle\, and\, Glass}) > P_{{Thirst}}({\rm\it Glass})
\end{equation}
for the notion of {{\it Thirst}, if we consider the presence of `liquid' as a valid sign for the presence of the notion {\it Thirst}.

So, 
we have proven the presence of `emergence' 
for 
the concept of {\it Love},  for the combination of concepts 
{\it Mother and Child}, and we have done it also for  
the concept of {\it Thirst}, for the combination of concepts 
{\it Bottle and Glass}, 
in a quantitative way, by making use of the Google Images search 
tool 
on the World-Wide Web. More specifically, if {\it Mother} and {\it Child} appear together in an image this increases the probability that also an embrace as a sign of {\it Love} appears in that picture, as compared to such an embrace appearing as a sign of {\it Love} if pictures are considered containing {\it Mother} or containing {\it Child}. Hence it shows that combining {\it Mother} and {\it Child} within an image increases the probability of the emergence of {\it Love} in this image, and compared to such an emergence taking place in 
images 
containing 
{\it Mother} or 
images 
containing 
only 
{\it Child}. A similar phenomenon of `increased emergence' 
for 
{\it Thirst} takes place when {\it Bottle} and {\it Glass} appear both in an image as compared to the emergence of {\it Thirst} for an image containing 
only 
{\it Bottle} or an image containing 
only 
{\it Glass}. This `increased 
emergence', called `overextension',  
can be modelled by the phenomenon of `interference'. 
Indeed, the quantum theoretical models that were developed and shown to 
successfully 
account for 
this type of overextension 
in situations of cognitive science, such as the Linda problem and the pet-fish problem \cite{aerts2009a,aertsgaborasozzo2013,sozzo2014,aertsetal2017a,aertssozzoveloz2015a,aertssozzoveloz2015b,aertssozzoveloz2015c}, 
which 
can equally be used 
in the present visual perception situation, 
precisely used interference effects as the main quantum effect to explain the deviation observed with respect to classical probabilities.

\section{Quantum entanglement in visual perception}

Encouraged by the results exposed in the foregoing sections, namely the violation of the conjunction inequalities leading to the identification of overextension in the probabilities calculated from data connected to images on the World-Wide Web, I was eager to explore more of the quantum nature of an image. The question that came to mind is `whether it 
were 
possible to identify the presence of quantum entanglement between different aspects of an image'. The violation of the conjunction inequalities made it possible to identify the presence of Linda 
and pet-fish problem type situations in visual perception. Could we use the same method of calculating probabilities 
using Google Images to violate a Bell type inequality and demonstrate the presence of quantum entanglement? This indeed turned out to be the case.

Making use of the example 
of the conceptual combination 
{\it The Animal Acts}, combining the two concepts {\it Animal} and {\it Act}, and different exemplars of this combination, 
then using Google Images, and the extraction of 
the 
corresponding probabilities as relative frequencies of appearances of the combinations of the exemplars, 
I was 
able to violate in a significant way the Clauser Horne Shimony Holt (CHSH) version of Bell's inequalities \cite{clauser1969}, in this way showing the presence of quantum entanglement in visual perception. In the following 
I will 
explain in detail how this violation comes about. 
But before that, it is worth observing 
that a violation of the CHSH inequality was shown to 
occur 
for the same combination of concepts {\it The Animal Acts} and the same set of exemplars than the one we will consider here, 
in 
cognitive experiments 
with 
human subjects \cite{aertssozzo2011}. The CHSH inequality is violated in case 
a specific term in it (se below) 
is smaller than 
$-2$ 
or bigger than 
$2$, 
and the numerical violation obtained in \cite{aertssozzo2011} is equal to 
$2.4197$, 
while 
as we are going to see 
I 
will 
obtain a numerical violation equal to 
$2.4107$. 
That these violations are so numerically close, while the probabilities come about from completely different types of experiments, 
can be taken as a sign of 
the robustness of the presence of the entanglement, in the first case 
when the combined concepts are elements of the human language and in the second case when they are the visual elements of pictures. 

I will 
explain now 
how 
this violation 
comes about. To calculate the term $E(A',B')+E(A,B')+E(A',B)-E(A,B)$ that appears in the CHSH inequality
\begin{eqnarray} \label{chshinequality}
-2 \le E(A',B')+E(A,B')+E(A',B)-E(A,B) \le 2
\end{eqnarray}
we need to consider four experiments $e(A,B)$, $e(A,B')$, $e(A',B)$ and $e(A',B')$, 
each of them 
consisting in a coincidence experiment. For example, $e(A,B)$ is the experiment consisting in jointly performing the measurements of concepts $A$ and $B$.
The quantities $E(A,B)$, $E(A,B')$, $E(A',B)$ and $E(A',B')$,
which are averages, 
will 
then 
be calculated from the data gathered by 
the experiments $e(A,B)$, $e(A,B')$, $e(A',B)$ and $e(A',B')$, 
respectively, as 
it will be explained 
in the following.

I will 
first explain
what the experiment $e(A,B)$ is. 
Consider 
for the concept  {\it Animal} the 
two exemplars {\it Horse} and {\it Bear}, 
as outcomes, 
and for the concept {\it Acts} 
the 
two exemplars {\it Growls} and {\it Whinnies}, 
as outcomes. 
The four combinations {\it The Horse Growls}, {\it The Horse Whinnies}, {\it The Bear Growls} and {\it The Bear Whinnies}, where each of 
them 
is an exemplar of {\it The Animal Acts}, 
constitute 
the four 
outcomes of experiment $e(A, B)$, where $A$ and $B$ are jointly measured. 
To obtain the probabilities associated with these four outcomes, 
I proceeded 
 as follows.
In Google Images 
I made 
a search on `horse growls', 
entering the two words one after the other in the Google bar,  
and 
saw 
that 400 images 
were 
shown to 
me 
by one hit of Google Images. 
I then carefully 
inspected 
each of the 400 images and 
counted 
the number of them on which 
I saw 
a `horse that growls'. This of course contains some subjective appreciation of the images, because 
I only had access to 
visual information. Doing so, 
I 
concluded that of the 400 images there were 4 possibly showing 
`a horse that is growling' (instead of whinnying, which is of course the most common noise for a horse to make, or instead of the horse being shown making no sound at all, 
or 
instead of the image showing no horse at all). 

I then used 
the same method as in the foregoing sections to come to a probability, namely we divide the number of 
images showing a horse that growls by the total number of images inspected. This gives 
$P(A_1,B_1)={4\over 400}=0.01$. 
The second sub experiment consists of making a search on Google Images for `horse whinnies'. This time 391 images showed up as search results. A careful verification of each image gave 
the count of 39 of them showing
`a horse that whinnies'. Hence, the probability that 
I was able to 
derive for this sub experiment is 
$P(A_1,B_2)={39\over 391}\approx 0.0997$. 
Let 
me 
mention that 
I 
use $A_1$ to indicate the exemplar {\it Horse} for {\it Animal} and $A_2$ to indicate the exemplar {\it Bear} for {\it Animal}. Analogously $B_1$ indicates {\it Growls} for {\it Acts} and $B_2$ indicates {\it Whinnies} for {\it Acts}. The third sub experiment of $e(A,B)$ consists of making a search on Google Images for `bear growls'. This time 380 images appeared after the search. 
I could count 
142 of them showing
`a bear that growls'. Hence this 
led me
to 
$P(A_2,B_1)={142\over 380}\approx 0.3737$. 
The fourth sub experiment of $e(A,B)$ consists of making a search on Google Images for `bear whinnies', and 389 images appeared after the search. 
I 
counted $2$ of them 
able to be interpreted as 
`a bear that whinnies'. Hence this 
lead me 
to 
$P(A_2,B_2)={2\over 389}\approx 0.0050$.

A comment is at place now to specify the way these counts have been done. A first remark is that they turn out to be rather objective, what would perhaps not be expected at first sight. Indeed, 
I asked several friends to also look over the pictures and 
give their estimations. It turned out that a rather good agreement on the numbers of 
pictures 
where one  imagines `the horse to be growling', or `the horse to be whinnying' or `the bear to be growling', or `the bear to be whinnying' was obtained.  And this was also the case for all the other appearances of one of the considered exemplars of {\it The Animal Acts} in the experiments we will consider 
in the following. 
What is however more important to mention is the following. That 
a picture 
contained or 
not 
`a horse whinnying' or `a bear growling' is really quite objective, because these are two common acts of a horse and a bear. It is more difficult to make a decision 
whether a specific image is about the less common happenings of `the horse growling' or 'the bear whinnying'.  
With respect to these non common situations, and they appear for each of the sub experiments, we decided to count an image as positive for it, whenever it was possible to imagine the not so common acts of said 
animals as pictured in the images.  
Even such an 
easygoing 
attitude 
with respect to the 
uncommon 
exemplars led to small numbers of their appearances, because it was very easy in each case to definitely decide that the 
uncommon 
exemplar was `not' pictured. That this is a valuable way to proceed stems from the following fact. Each time such an image is counted and the interpretation would be too 
wide, 
e.g. one imagines that the horse on the image could be growling while if one would have been present when the 
picture was taken 
one would have heard this not to be the case, this will only 
affect 
the term in the CHSH inequality 
in a way that the latter will be 
`less easily violated'. 
In 
other words, errors on 
measurements for the less common exemplars, due to a too broad interpretation, 
 always push the term in the CHSH inequality 
towards 
the direction of a value closer to be contained in the interval $[-2,+2]$ where no violation occurs. 

This aspect of the CHSH inequality, its enormous robustness for errors, is also the one 
explaining 
why the violation measured with coupled spins in quantum theory was such an important happening, indicating the presence of quantum entanglement in the physical micro-world \cite{aspectgrangierroger1981}. Indeed, also 
for spins it was 
true that errors due to imperfections of the experiments would only push the CHSH term in the inequality further away from 
its 
violation. 
I made screen pictures of all of the images indicating the ones that were withheld for `horse growling', and also screen pictures of all the other Google Images search results and how I came to the decision to count the ones that confirm the specific exemplar of {\it The Animal Acts}.
Hence, 
any reader can receive these images on request to verify how the decisions to withhold images was made.

For this first sub experiment $e(A,B)$ we get four probabilities 
${\cal P}(A_1,B_1)$, ${\cal P}(A_1,B_2)$, ${\cal P}(A_2,B_1)$ and ${\cal P}(A_2,B_2)$, 
respectively connected to the chance to find a `horse that growls', a `horse that whinnies', a `bear that growls'
and 
a `bear that whinnies', in the collection of Google Images 
obtained 
when the searches are made as explained above. 
However, to 
formulate the CHSH inequality 
we need 
to further normalize these quantities. 
More specifically, we need estimations of `what is the probability that one of the four will appear if we know that at least one of them is appearing'. We can 
obtain 
these renormalised probabilities 
by dividing each of the now calculated probability 
${\cal P}$  
by their 
sum (note that all decimal numbers indicated below are approximate numbers, so the equality symbols should be more precisely replaced by approximate equality  symbols): 
\begin{equation}
{\cal S}(A,B)={\cal P}(A_1,B_1)+{\cal P}(A_1,B_2)+{\cal P}(A_2,B_1)+{\cal P}(A_2,B_2)=0.4884
\end{equation}
This gives: 
\begin{eqnarray}
&&P(A_1,B_1)={{\cal P}(A_1,B_1) \over {\cal S}(A,B)}=0.0205 \quad\quad P(A_1,B_2)={{\cal P}(A_1,B_2) \over {\cal S}(A,B)}=0.2042 \\
&&P(A_2,B_1)={{\cal P}(A_2,B_1) \over {\cal S}(A,B)}=0.7651 \quad\quad P(A_2,B_2)={{\cal P}(A_2,B_2) \over {\cal S}(A,B)}=0.0103 \\
\end{eqnarray}
The 
obtained 
expectation value $E(A,B)$ 
is then 
\begin{eqnarray}
E(A,B)=P(A_1,B_1)-P(A_1,B_2)-P(A_2,B_1)+P(A_2,B_1)=-0.9385
\end{eqnarray}
The idea is that 
the 
choice for {\it Animal} which is {\it Horse} is 
given the value $1$, 
while the choice for {\it Animal} 
which 
is {\it Bear} is 
given the value $-1$. 
Similarly, the choice for {\it Acts} which is {\it Growls} is 
given the value $1$, 
and the choice for {\it Acts} which is {\it Whinnies} is 
given the value $-1$. 
Then, combining these values, we obtain that 
the choice 
{\it The Horse Growls} is 
associated with the value $1$, obtained by multiplying the  value $1$ for {\it Horse} with the value $1$ for {\it Growls}. Similarly, 
the choice 
{\it The Horse Whinnies} is 
$-1$ (multiplying $1$ with $-1$),  
the choice 
{\it The Bear Growls} is 
$-1$ (multiplying $-1$ with $1$),  
and the choice 
{\it The Bear Whinnies} is 
$1$ (multiplying $-1$ with $-1$).  
Then, 
$E(A,B)$ is 
the `expected value' given  
by 
the probabilities 
$P(A_1,B_1)$, $P(A_1,B_2)$, $P(A_2,B_1)$ and $P(A_2,B_1)$ 
of each of these values. Hence $E(A,B)=-0.9385$ means that there is a strong anti-correlation, and indeed, {\it Horse} anti-correlates with {\it Growls} and {\it Bear} 
anti-correlates 
with {\it Whinnies}.

To define the three remaining experiments $e(A,B')$, $e(A',B)$ and $e(A',B')$ 
that are 
needed to calculate the CHSH inequality term, we consider two different exemplars for {\it Animal} as well as for {\it Acts}, namely {\it Tiger} and {\it Cat} and {\it Snorts} and {\it Meows}. The three experiments are now defined as follows, $e(A,B')$ is the experiment where the old exemplars for {\it Animal}, {\it Horse} and {\it Bear}, are combined with the new exemplars for {\it Acts}, {\it Snorts} and {\it Meows}, $e(A',B)$ is the experiment where the new exemplars for {\it Animal}, {\it Tiger} and {\it Cat}, are combined with the old exemplars for {\it Acts}, {\it Growls} and {\it Whinnies}, and $e(A',B')$ is the experiments where for both {\it Animal} and {\it Acts} the new exemplars are combined, hence {\it Tiger} and {\it Cat} with {\it Snorts} and {\it Meows}.

So, for the experiment $e(A,B')$, I did 
Google Images searches  for `horse snorts', delivering 407 images of which 41 show a snorting 
horse; 
for `horse meows', delivering 403 images of which 4 show a meowing 
horse; 
for `bear snorts', delivering 385 images of which 23 show a snorting 
bear; and 
for `bear meows', delivering 405 images of which 5 show a meowing bear. This gives  
${\cal P}(A_1,B'_1)=0.1007$, ${\cal P}(A_1, B'_2)=0.0099$, ${\cal P}(A_2,B'_1)=0.0597$ and ${\cal P}(A_2,B'_2)=0.0123$, 
and hence 
\begin{eqnarray}
&&{\cal S}(A,B')={\cal P}(A_1,B'_1)+{\cal P}(A_1,B'_2)+{\cal P}(A_2,B'_1)+{\cal P}(A_2,B'_2)=0.1827  \\
&&P(A_1,B'_1)={{\cal P}(A_1,B'_1) \over {\cal S}(A,B')}=0.5512 \quad\quad P(A_1,B'_2)={{\cal P}(A_1,B'_2) \over {\cal S}(A,B')}=0.0543 \\
&&P(A_2,B'_1)={{\cal P}(A_2,B'_1) \over {\cal S}(A,B')}=0.3269 \quad\quad P(A_2,B'_2)={{\cal P}(A_2,B'_2) \over {\cal S}(A,B')}=0.0676 \\
&&E(A,B')=P(A_1,B'_1)-P(A_1,B'_2)-P(A_2,B'_1)+P(A_2,B'_1)=0.2376
\end{eqnarray}
For 
the experiment $e(A',B)$, 
I did 
Google Images searches for `tiger growls', delivering 399 images of which 219 show a growling 
tiger; 
for `tiger whinnies', delivering 402 images of which 3 show a whinnying 
tiger; 
for `cat growls', delivering 405 images of which 78 show a growling 
cat; 
and for `cat whinnies', delivering 402 images of which 1 shows a whinnying cat. This gives 
${\cal P}(A'_1,B_1)=0.5489$, ${\cal P}(A'_1, B_2)=0.0075$, ${\cal P}(A'_2,B_1)=0.1926$ and ${\cal P}(A'_2,B_2)=0.0025$, 
and hence 
\begin{eqnarray}
&&{\cal S}(A',B)={\cal P}(A'_1,B_1)+{\cal P}(A'_1,B_2)+{\cal P}(A'_2,B_1)+{\cal P}(A'_2,B_2)=0.7514  \\
&&P(A'_1,B_1)={{\cal P}(A'_1,B_1) \over {\cal S}(A',B)}=0.7305 \quad\quad P(A'_1,B_2)={{\cal P}(A'_1,B_2) \over {\cal S}(A',B)}=0.0099 \\
&&P(A'_2,B_1)={{\cal P}(A'_2,B_1) \over {\cal S}(A',B)}=0.2563 \quad\quad P(A'_2,B_2)={{\cal P}(A'_2,B_2) \over {\cal S}(A',B)}=0.0033 \\
&&E(A',B)=P(A'_1,B_1)-P(A'_1,B_2)-P(A'_2,B_1)+P(A'_2,B_1)=0.4675
\end{eqnarray}
Finally, for 
the experiment $e(A',B')$, 
I did 
Google Images searches for `tiger snorts', delivering 400 images of which 30 show a snorting 
tiger;  
for `tiger meows', delivering 403 images of which 10 show a meowing 
tiger; 
for `cat snorts', delivering 392 images of which 15 show a snorting 
cat; 
and for `cat meows', delivering 399 images of which 161 show a meowing cat. This gives 
${\cal P}(A'_1,B'_1)=0.075$, ${\cal P}(A'_1, B'_2)=0.0248$, ${\cal P}(A'_2,B'_1)=0.0383$ and ${\cal P}(A'_2,B'_2)=0.4035$, 
and hence
\begin{eqnarray}
&&{\cal S}(A',B')={\cal P}(A'_1,B'_1)+{\cal P}(A'_1,B'_2)+{\cal P}(A'_2,B'_1)+{\cal P}(A'_2,B'_2)=0.5416  \\
&&P(A'_1,B'_1)={{\cal P}(A'_1,B'_1) \over {\cal S}(A',B')}=0.1385 \quad\quad  P(A'_1,B'_2)={{\cal P}(A'_1,B'_2) \over {\cal S}(A',B')}=0.0458 \\
&&P(A'_2,B'_1)={{\cal P}(A'_2,B'_1) \over {\cal S}(A',B')}=0.0707 \quad\quad  P(A'_2,B'_2)={{\cal P}(A'_2,B'_2) \over {\cal S}(A',B')}=0.7450 \\
&&E(A',B')=P(A'_1,B'_1)-P(A'_1,B'_2)-P(A'_2,B'_1)+P(A'_2,B'_1)=0.7671
\end{eqnarray}
We have now all available measurements and their corresponding calculations of probabilities and expectation values to calculate the term of the CHSH inequality, and this gives
\begin{eqnarray}
E(A',B')+E(A,B')+E(A',B)-E(A,B)=2.4107
\end{eqnarray}
Hence, 
the CHSH inequality (\ref{chshinequality}) is manifestly violated.

\section{Entanglement and meaning connection}
The violation of the conjunction inequalities indicates the presence of `emergent meaning'. Linda fits  this emergent meaning well for the conjunction `bank teller' and `active on the feminist movement' and that is why systematically the 
so-called conjunction fallacy 
is made when the 
corresponding experiment is done. Guppy fits the emergent meaning well for the conjunction 
of 
`pet' and `fish', and tomato resonates with the emergent meaning of `fruits' and `vegetables'. We can also understand the violation of Bell's inequalities and hence the identification of the presence of entanglement by reflecting about the meaning content of the considered situation. 
Let me 
illustrate this also by analysing some of the images that showed up when 
I 
performed the Google Images searches described in detail in the foregoing section and leading to the probabilities 
violating 
the CHSH version of Bell's inequalities. 

In a search for 
`horse whinnies' the typical image that showed up was like 
the one 
illustrated in Figure~\ref{horse-whinnies}, where one can see a horse whinny with its mouth being shaped in 
the typical form. 
\begin{figure}[H]
\begin{center}
\includegraphics[width=8cm]{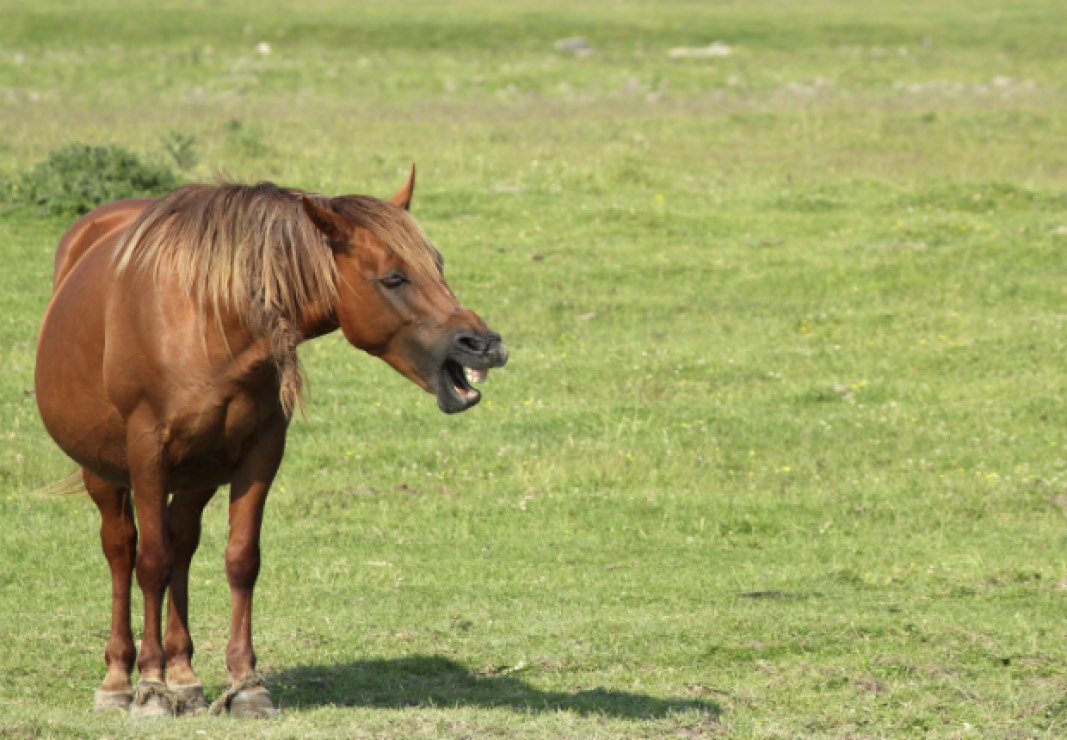}
\caption{A typical 
image 
showing up when a search 
for 
`horse whinnies' is being made in Google Images
\label{horse-whinnies}}
\end{center}
\end{figure}
\noindent
Also for `bear growls', `tiger growls' and `cat meows' the typical images 
showing these behaviors will 
appear, 
and 
the reader is invited to try this out in Google 
Images. Certainly, 
`bear growls' gives rise to an impressive amount of pictures of heavily growling bears. However, even for a search of `cat meows' also 
images 
will always 
appear in the search that do not show a `meowing cat'. For example, the image in Figure~\ref{cat-meows} is the 19th image after 18 meowing cats.
\begin{figure}[H]
\begin{center}
\includegraphics[width=8cm]{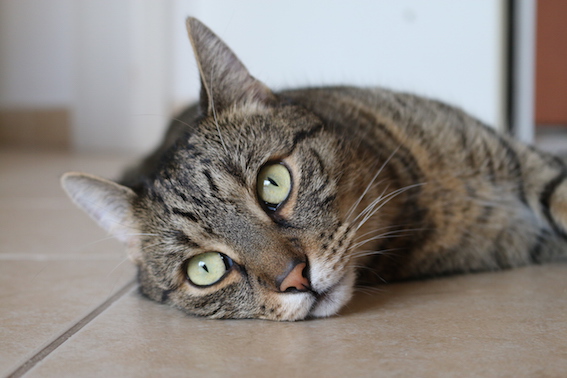}
\caption{The 19th image appearing 
in 
a search of `cat meows' in Google Images is no longer a cat that meows.
\label{cat-meows}}
\end{center}
\end{figure}
\noindent
 In the first experiment $e(A,B)$ these two exemplars, `the horse whinnies' and `the bear 
growls', 
are the dominant ones in 
terms of 
appearance, while the other two 
exemplars, 
`the horse growls' and `the bear 
whinnies', 
are very uncommon, and we have counted some of them mostly because some images might 
lend themselves to be interpreted in this way. 
It is also this rather strong variation of dominance of two of the exemplars versus 
the 
uncommon appearance of two of the other exemplars  
that brings 
the expectation value $E(A,B)$ 
to be close to $-1$, which is 
an important factor for the CHSH inequalities to be able to be violated. It is of course the `meaning connection' between {\it Animal} and {\it Acts} which is at the origin of this phenomenon of dominance of two of the exemplars 
and the 
uncommon appearance of the other two. 
However, 
connection in meaning is not enough to violate the CHSH inequalities, we can see this easily by considering a special case, where we take $A'=A$ and $B'=B$. Since then $E(A',B')=E(A,B')=E(A',B)=E(A,B)$ the term in the CHSH inequality becomes
equal to $2E(A,B)$, 
and since $-1 \le E(A,B) \le +1$, the inequality will never be violated.

This special case shows that to violate the 
inequalities, 
a meaning connection is necessary, but the source 
of it, in our case the conceptual combination {\it The Animal Acts}, 
needs also to be able to respond to different experiments in different ways. This 
`creation aspect' of being ble to `respond 
to different experiments in different ways' is also very typical of the potentiality contained in a quantum state \cite{aerts2009b,aerts2010a,aerts2010b}. This potentiality is carried by the 
very 
conceptual nature of 
abstract entities 
such as {\it Animal} and 
{\it Acts}: 
they can realise the 
more concrete 
states of {\it Horse}, {\it Bear}, {\it Tiger} and {\it Cat} for the case of {\it Animal}, and {\it Growls}, {\it Whinnies}, {\it Snorts} and {\it Meows} for the case of {\it Acts}, each time 
in a different way (as in a process of symmetry breaking), being only guided by the 
existing meaning connection.

\section{Conclusion}
When I worked on the experiments about the after 
image, during my Master studies in visual arts 
\cite{aertsarguelles2017b}, I would not have imagined 
 that soon after I would be studying the violation of 
 conjunction and Bell's inequalities  due to the presence of quantum superposition and quantum entanglement
in visual perception. 
But 
I know now that additionally to my quest for the heart of an image, guided by my long lasting fascination for the content and meaning carried by an image, 
a 
long-running circle was 
coming to its closure. I remember how my father would ask me and my friends, playing as children in the garden not far away from where he was working, how we would think about an `tree house' 
whether we believed it to be rather a `building' or a `dwelling' or somewhat 
both.
I also remember him asking us about `mint' and whether we considered it to be a plant or food or rather both. He also told us about the `pet-fish', and how `guppy' and `goldfish' behaved weirdly with respect to it, and how he was investigating whether this behaviour could be understood by making use of quantum theory. I remember how these 
intermezzi 
in our playing would also lead to listening to a presentation of the 
double-slit 
experiment and the weird behaviour of quantum particles. 
However, 
I would never have fathomed then that my studies in visual art and my fascination for the ontology of an image would come to be connected 
in such a 
meaningful way 
to what my father was presenting to me and my friends in 
those days. 
But so it 
was.

\end{document}